\def\year{2022}\relax
\documentclass[letterpaper]{article} 
\usepackage{aaai22}  
\usepackage{times}  
\usepackage{helvet}  
\usepackage{courier}  
\usepackage[hyphens]{url}  
\usepackage{graphicx} 
\usepackage{natbib}  
\usepackage{caption} 
\DeclareCaptionStyle{ruled}{labelfont=normalfont,labelsep=colon,strut=off} 
\usepackage{multirow}
\usepackage{multicol}
\usepackage{arydshln}

\usepackage{algorithm}
\usepackage{algorithmic}

\usepackage{newfloat}
\usepackage{listings}
\usepackage[utf8]{inputenc} 
\usepackage[T1]{fontenc}    
\usepackage{url}            
\usepackage{booktabs}       
\usepackage{amsfonts}       
\usepackage{nicefrac}       
\usepackage{microtype}      
\usepackage{bm}
\usepackage{amsmath}
\usepackage{epstopdf}
\usepackage{graphicx}
\usepackage{wrapfig}
\usepackage{color}
\usepackage{subfigure}
\usepackage{extarrows}
\usepackage{amsmath}
\usepackage{ntheorem}
\usepackage{wrapfig}

\usepackage{xcolor}

\newtheorem{proposition}{Proposition}

\newtheorem{definition}{Definition}

\newtheorem*{proof}{Proof}

\floatstyle{ruled}
\newfloat{listing}{tb}{lst}{}
\floatname{listing}{Listing}
\title{Geometry Interaction Knowledge Graph Embeddings}
\author{
Zongsheng Cao \textsuperscript{\rm 1,2},
   Qianqian Xu \textsuperscript{\rm 3,*},
   Zhiyong Yang \textsuperscript{\rm 4},
   Xiaochun Cao \textsuperscript{\rm 1,2},
   Qingming Huang \textsuperscript{\rm 3,4,5,6,*}
}
\affiliations{
\textsuperscript{\rm 1} State Key Laboratory of Information Security, Institute of Information Engineering, CAS, Beijing, China \\
   \textsuperscript{\rm 2} School of Cyber Security, University of Chinese Academy of Sciences, Beijing, China \\
   \textsuperscript{\rm 3} Key Laboratory of Intelligent Information Processing, Institute of Computing Technology, CAS, Beijing, China \\
   \textsuperscript{\rm 4} School of Computer Science and Technology, University of Chinese Academy of Sciences, Beijing, China \\
   \textsuperscript{\rm 5} Key Laboratory of Big Data Mining and Knowledge Management, Chinese Academy of Sciences, Beijing, China \\
   \textsuperscript{\rm 6} Peng Cheng Laboratory, Shenzhen, China \\
$\{$caozongsheng,caoxiaochun$\}$@iie.ac.cn, xuqianqian@ict.ac.cn, yangzhiyong21@ucas.ac.cn, qmhuang@ucas.ac.cn
}
\usepackage{bibentry}

\begin{document}

\maketitle

\begin{abstract}
Knowledge graph (KG) embeddings have shown great power in learning representations of entities and relations for link prediction tasks. Previous work usually embeds KGs into a single geometric space such as  Euclidean space (zero curved),  hyperbolic space (negatively curved) or  hyperspherical space (positively curved) to maintain their specific geometric structures (e.g., chain, hierarchy and ring structures). However, the topological structure of KGs appears to be complicated, since it may contain multiple types of geometric structures simultaneously. Therefore, embedding KGs in
a single space, no matter the Euclidean space, hyperbolic space or hyperspheric space, cannot capture the complex structures of KGs accurately. To overcome this challenge, we propose Geometry Interaction knowledge graph Embeddings (GIE), which learns spatial structures interactively between the Euclidean, hyperbolic and hyperspherical spaces. Theoretically, our proposed GIE can capture a richer set of relational information, model key inference patterns, and enable expressive semantic matching across entities. Experimental results on three well-established knowledge graph completion benchmarks show that our GIE achieves the state-of-the-art performance with fewer parameters.
\end{abstract}

\section{Introduction}

Knowledge Graphs (KGs) attract more and more attentions in the past years, which 
play an important role in many semantic applications (e.g., question answering \cite{DBLP:conf/www/DiefenbachSM18,DBLP:conf/eacl/LascaridesLGKCE17}, semantic search \cite{DBLP:conf/emnlp/BerantCFL13,DBLP:conf/acl/BerantL14}, recommender systems \cite{DBLP:conf/sigir/Gong0WFP0Y20,DBLP:journals/corr/abs-2003-00911},
 and
dialogue generation \cite{DBLP:conf/eacl/LascaridesLGKCE17,DBLP:conf/acl/HeBEL17}). KGs
can not only represent vast amounts of information available in the world, but also enable powerful relational reasoning. However, real-world KGs are usually incomplete, which restricts the  development of  downstream tasks above. To address such an issue, Knowledge
Graph Embedding (KGE) emerges as an effective solution to
predict missing links by the low-dimensional representations of
entities and relations.

In the past decades,  researchers propose various  KGE methods which  achieve great success. As a branch of KGE methods, learning in the Euclidean  space $\mathbb{E}$ and ComplEx space $\mathbb{C}$ 
is  highly effective, 
 which has been  demonstrated by methods such as DistMult \cite{yang2014embedding}, ComplEx \cite{trouillon2016complex} and
TuckER \cite{balavzevic2019tucker} (i.e., bilinear models). These methods   can infer new relational triplets and capture  chain structure with the scoring function based on the intuitive Euclidean distance.
However, learning embeddings in Euclidean space are weak to capture complex structures (e.g., hierarchy and ring structures) \cite{2019Multi}.

\begin{figure*}
	\centering
\includegraphics[width=6.3in]{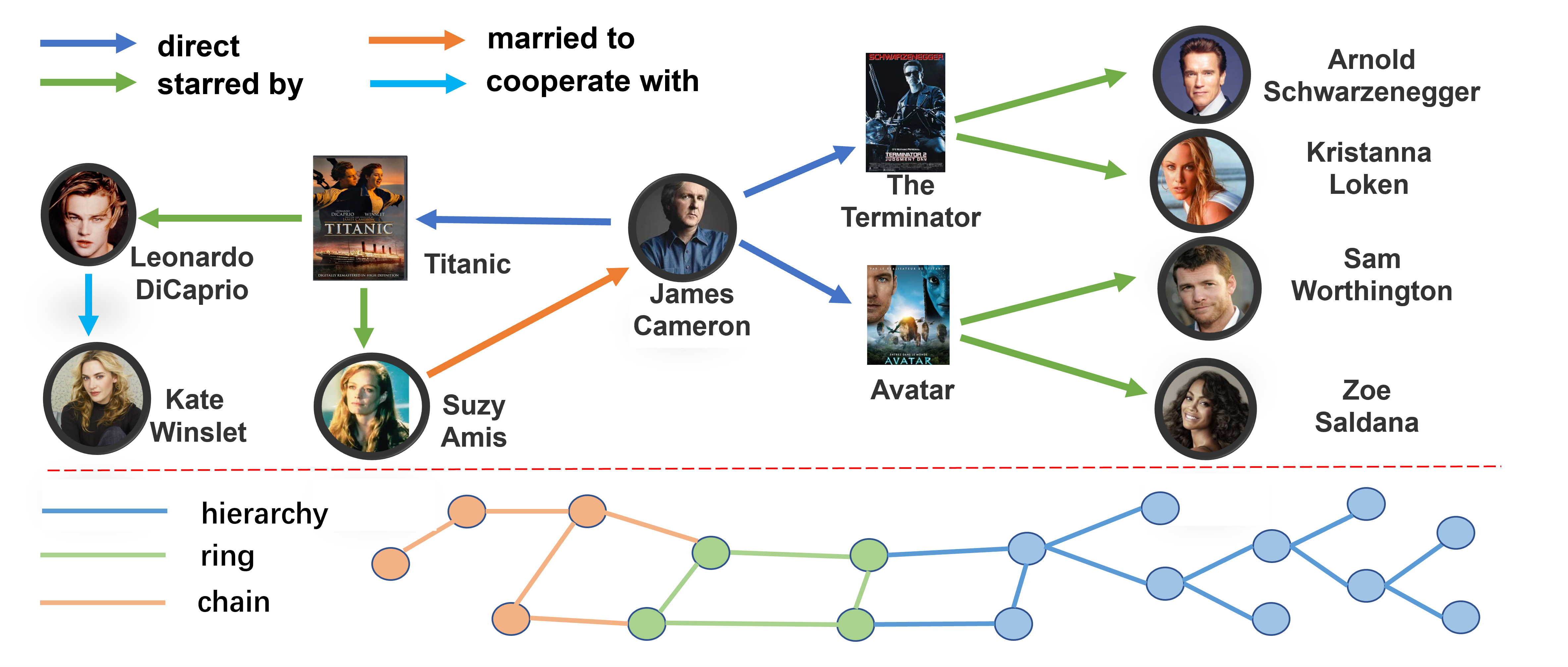}
  \caption{ An illustration of the structure in  knowledge graphs. The first part is a knowledge graph about the film and characters directed by James Cameron. The second part is the different spatial structure formed by different types of data such as hierarchical structure,  ring structure and chain structure. Orange entities of the KG show the chain structure which can be easily modeled in an Euclidean space;
  green entities of the KG present ring structure which can be depicted  in a hypersphere space;
  blue entities of the KG  exhibit hierarchical structures which can be intuitively captured in a hyperbolic
space.}\label{KG}
\end{figure*}

As another branch of KGE methods, learning KG embeddings in the non-Euclidean space (e.g., hyperspheric space $\mathbb{S}$ and hyperbolic space $\mathbb{H}$) has received wide attention in recent years. 
Drawing on the geometric properties of the sphere,  
ManifoldE \cite{DBLP:conf/ijcai/0005HZ16} embeds entities into the manifold-based space such as hypersphere space, which is beneficial to improve the precision of knowledge embedding and model the ring structure.
 Due to the capability of hyperbolic
geometry in modeling  hierarchical structure,  methods such as MuRP \cite{2019Multi} and \textsc{RotH} \cite{chami2020low} can outperform the methods based on Euclidean space and Complex  space in  knowledge graphs of hierarchical structures. 
However, note that most knowledge graphs  embrace complicated  structures as shown in Figure \ref{KG}, it implies that these methods perform a bit poorly when modeling the knowledge graph with hybrid structures   \cite{2019Multi} (e.g., ring, hierarchy  and chain structures).
 Recently, \cite{wang2021mixed}  develops a  mixed-curvature multi-relational graph neural network for knowledge graph completion, which is constructed by a Riemannian product manifold between different spaces. However, it does not delve into the interaction of the various spaces  and the heterogeneity of different spaces will limit its pattern reasoning ability between spaces.

From the analysis above, we can see learning knowledge graph embeddings without geometry interaction will inevitably result in sub-optimal results. The reason lies in that it cannot accurately capture the complex structural features of the entity location space, and the transfer of inference rules in different spaces will be affected or even invalid.  
Inspired by this,
 to capture a richer set of structural information and model key inference patterns for KG with complex structures,
  we attempt to embed the knowledge graph into Euclidean,  hyperbolic and hyperspherical spaces jointly  by geometry interaction, hoping that it can  integrate  the reasoning ability in different spaces. 

In this paper, we take a step beyond
the previous work,  and
propose a novel method called Geometry Interaction Knowledge Graph Embeddings (\textbf{GIE}). 
GIE can not only
exploit hyperbolic, hyperspherical and Euclidean geometries simultaneously   to learn the reliable structure in knowledge graphs,
but also 
model inference rules. Specifically, 
we model the relations by rich geometric transformations based on \emph{special orthogonal group} and \emph{special Euclidean group} to utilize key inference patterns, enabling rich and expressive semantic
matching between entities. Next, to approximate the complex
structure  guided by heterogeneous spaces,
 we  exploit a novel geometry interaction mechanism,  which is designed to capture the geometric message for
complex structures in KGs.  
In this way, GIE can  adaptively form
the reliable spatial structure for knowledge graph embeddings via integrating the interactive geometric message.
Moreover,
experiments  demonstrate that GIE can significantly outperform the state-of-the-art methods.

Our contributions are summarized as follows:
    \textbf{1)} To the best of our knowledge, we are the first to exploit knowledge graph embeddings with geometry interaction, which adaptively learns reliable spatial structures  and is equipped with the ability to transmit inference rules in complex structure of KGs.
   \textbf{2)} We present a novel method called Geometry Interaction Knowledge Graph Embeddings (GIE), which embraces rich and expressive semantic matching between entities and  satisfies  the key  of relational representation learning.
(i.e., modeling symmetry, anti-symmetry, inversion, composition, hierarchy, intersection and mutual exclusion patterns).
   \textbf{3)} Our model requires fewer parameters and saves about $40\%$ parameters than SOTA models, while achieving the state-of-the-art performance 
on three well-established knowledge graph completion benchmarks (WN18RR, FB15K237 and YAGO3-10).

\section{Related Work}
The key issue of  knowledge graph embeddings is to learn low-dimensional distributed embedding of entities and relations. Previous work has proposed many KGE methods based on  geometric properties of different embedding spaces. These methods mainly use Euclidean space (e.g., the vector space, matrix space and tensor space), while some methods are based on other kinds of spaces such as complex space and hyperbolic space. Moreover, there are some methods based on deep neural networks utilized as well.

\noindent\textbf{Euclidean embeddings.}  In recent years, researchers have conducted a lot of research based on Euclidean embeddings for KG representation learning. These methods include
translation approaches popularized by TransE \cite{bordes2013translating}, which  models the relation as the translation between head and tail entities, i.e., $head \ entity + relation \approx tail \ entity$.
Then there are many variations  based on TransE.
 TransH \cite{wang2014knowledge}, TransR \cite{lin2015learning}, TransD \cite{ji2015knowledge} use different projection vectors for entities and relations, separately. TorusE \cite{ebisu2018toruse:} adopts distance function in a compact Lie group to avoid forcing embeddings to be on a sphere. TransG \cite{xiao2016transg} and KG2E \cite{He2015Learning} further inject probabilistic principles into this framework
by considering Bayesian nonparametric Gaussian mixture model and show
better accuracy and scalability.
  And there are also tensor
factorization methods such as RRESCAL \cite{2011RESCAL}, DistMult \cite{yang2014embedding}. These methods are very simple and intuitive, but they cannot model the key logical properties (e.g., translations cannot model symmetry).

\noindent\textbf{Complex embeddings.} Instead of using a real-valued embedding, learning KG embeddings in the complex space has  attracted attention.
Some bilinear models such as ComplEx \cite{trouillon2016complex} and HolE \cite{nickel2015holographic}  measure plausibility by matching latent semantics
of entities and relations. 
Then, RotatE \cite{sun2019rotate:} models the relation as the rotation in two-dimensional space, thus realizing modeling symmetry/antisymmetry, inversion and composition patterns.
Inspired by RotatE, OTE \cite{tang-etal-2020-orthogonal}  extends the rotation from 2D to high-dimensional space by  relational transformation.
Moreover, QuatE \cite{zhang2019quaternion}  extends the
complex space into the quaternion space
with two rotating surfaces to enable rich semantic matching.
Following this, \cite{DBLP:conf/aaai/CaoX0CH21} extends the embedding space to the dual quaternion space, which models the interaction of relations and entities  by dual quaternion multiplication.
 However, complex space is essentially flat with Euclidean space since their curvatures of the space are zero.
 Therefore,  for the KGs with the hierarchical structure, these methods often suffer from large  space occupation and distortion embeddings.


\noindent\textbf{Non-Euclidean embeddings.} 
Noting the limitations of Euclidean space,
ManifoldE \cite{DBLP:conf/ijcai/0005HZ16} expands pointwise modeling in the translation based principle to manifoldwise space (e.g., hypersphere space), which benefits for modeling ring structures.
To capture the hierarchical structure in the knowledge graph, MuRP \cite{2019Multi} learns KG embeddings in hyperbolic space in order to target hierarchical data.
\cite{LCLR2019Mainfold} proposes learning embeddings in a product manifold with  a heuristic to estimate the sectional curvature of graph data. However, it ignores the interaction of the embedded spaces, which will affect the learning of spatial structure.
Then 
ROTH \cite{chami2020low} introduces the hyperbolic geometry on the basis of the rotation,
which can achieve low-dimensional knowledge graph embeddings. However, for non-hierarchical data, their performance will be limited.

\noindent\textbf{Deep neural networks.}
Neural network approaches such as \cite{DBLP:journals/corr/abs-2101-01669}, \cite{AS_GCN},\cite{nathani2019learning} and \cite{DBLP:conf/www/JinHL021}  play important roles in graph learning. Some deep neural networks have also been adopted for the KGE problem, e.g.,
Neural Tensor Network \cite{2013Reasoning} and ER-MLP \cite{2014Knowledge} are two representative
neural network based methodologies. Moreover, there are some models using neural networks to produce KGE with remarkable effects.  For instance, R-GCN \cite{schlichtkrull2018modeling}, ConvE \cite{dettmers2017convolutional}, ConvKB \cite{nguyen2017novel},  
  A2N \cite{bansal2019a2n} and HypER \cite{balavzevic2019hypernetwork}. Recently,  M$^2$GNN \cite{wang2021mixed} is proposed, which is a generic graph
neural network framework to embed multi-relational KG data for knowledge graph embeddings.
 A downside of these methods is that most of them are not interpretable.


\section{ Background and Preliminaries}

\textbf{Multi-relational link prediction.} The
multi-relational link prediction is an important task for knowledge graphs.
 Generally, given a set $\mathcal{E}$ of entities
and a set $\mathcal{R}$ of relations, a knowledge graph $\mathcal{G}\subseteq \mathcal{E}\times \mathcal{R}\times\mathcal{E}$ is a set of subject-predicate-object triples.
Noticing that the real-world knowledge graphs are often incomplete,
the goal of multi-relational link prediction is to complete the KG, i.e., to predict
true but unobserved triples based on the information in $\mathcal{G}$ with a score function.
Typically,  the score function is learned with the formula as $\phi:\mathcal{E}\times\mathcal{R}\times\mathcal{E}\rightarrow \mathbb{R}$, that assigns a score $s=\phi(\bm{h},\bm{r},\bm{t})$ to each triplet, indicating the strength of prediction that a
particular triplet corresponds to a true fact. Followed closely, a non-linearity, such as the logistic or sigmoid function, is
often used to convert the score to a predicted probability $p=\sigma(s)\in[0,1]$ of the triplet being true.

\noindent\textbf{Non-Euclidean Geometry}. Most of the manifolds in geometric space can be regarded as Riemannian manifolds $(\mathcal{M},g)$ of dimension $n$,
which define the internal product operation in the tangent space
$g_{\bm{x}}$ for each point $\bm{x}\in \mathcal{M}$: $\mathcal{T}_{\bm{x}}\mathcal{M}\times \mathcal{T}_{\bm{x}}\mathcal{M}\rightarrow \mathbb{R}$.
The tangent space $\mathcal{T}_{\bm{x}}\mathcal{M}$ of dimension $n$   contains all possible directions of the path in non-Euclidean space leaving from $\bm{x}$. 
Specifically, denote $g^E=\bm{I}_n$
 as the  Euclidean metric tensor and $c$ as the curvature of the space, then we can use the manifold $\mathbb{M}_{c}^{n}=\left\{\boldsymbol{x} \in \mathbb{R}^{n}:c\|\boldsymbol{x}\|<1\right\}$ to define non-Euclidean space $(\mathbb{M}_c^n,g^c)$, where the following Riemannian metric $g_{\bm{x}}^c=\lambda_{\bm{x}}^2g^{E}$ and $\lambda_{\bm{x}}=\frac{2}{1-c||\bm{x}||^2}$.

Note that
 the  curvature $c$  is positive for hyperspherical  space $\mathbb{S}$, negative for hyperbolic space $\mathbb{H}$, and zero
for Euclidean space $\mathbb{E}$.
Poincar$\acute{\text{e}}$ ball is  a popular adopted model in representation learning, which is commonly used in hyperbolic space and has basic mathematical operations
(e.g., addition, multiplication). Moreover,
it provides  closed-form expressions for many basic objects such as distance and angle \cite{DBLP:conf/nips/GaneaBH18}. The principled generalizations of basic operations in hyperspherical space are similar to operations in hyperbolic space, except that the curvature $c>0$.
Therefore, here we only introduce the operation of hyperbolic space, the operation of hyperspherical space can be obtained by analogy.

For each point $\bm{x}$ in the hyperbolic space, we have a tangent space $\mathcal{T}_{\bm{x}}\mathbb{H}_c^n$. 
We can use the  \emph{exponential} map $\exp^c_{\bm{x}}: \mathcal{T}_{\bm{x}}\mathbb{H}_c^n\rightarrow \mathbb{H}_c^n$ and the  \emph{logarithmic} map $\log_{\bm{x}}^c: \mathbb{H}_c^n
\rightarrow \mathcal{T}_{\bm{x}}\mathbb{H}_c^n$ to establish the connection between the
hyperbolic space and tangent space as follows:
\begin{equation}\label{two maps}
\begin{aligned}
&\exp _{\boldsymbol{x}}^{c}(\boldsymbol{v})=\boldsymbol{x} \oplus_{c}\left(\tanh \left(\sqrt{c} \frac{\lambda_{\boldsymbol{x}}\|\boldsymbol{v}\|}{2}\right) \frac{\boldsymbol{v}}{\sqrt{c}\|\boldsymbol{v}\|}\right),\\
&\log _{\boldsymbol{x}}^{c}(\boldsymbol{y})=\frac{2}{\sqrt{c} \lambda_{\boldsymbol{x}}} \tanh ^{-1}\left(\sqrt{c}\left\|-\boldsymbol{x} \oplus_{c} \boldsymbol{y}\right\|\right) \frac{-\boldsymbol{x} \oplus_{c} \boldsymbol{y}}{\left\|-\boldsymbol{x} \oplus_{c} \boldsymbol{y}\right\|},
\end{aligned}
\end{equation}
\normalsize{where} $\bm{x},\bm{y}\in \mathbb{H}_c^n$, $\bm{v}\in \mathcal{T}_{\bm{x}}\mathbb{H}_c^n$, 
 $||\cdot||$ denotes the Euclidean norm and $\oplus_c$ represents M$\ddot{o}$bius addition:
\begin{equation}
\boldsymbol{x} \oplus_{c} \boldsymbol{y}=\frac{\left(1+2 c\langle\boldsymbol{x}, \boldsymbol{y}\rangle+c\|\boldsymbol{y}\|^{2}\right) \boldsymbol{x}+\left(1-c\|\boldsymbol{x}\|^{2}\right) \boldsymbol{y}}{1+2 c\langle\boldsymbol{x}, \boldsymbol{y}\rangle+c^{2}\|\boldsymbol{x}\|^{2}\|\boldsymbol{y}\|^{2}}.
\end{equation}

The distance between two points $\bm{x},\bm{y}\in\mathbb{H}^d_c$ is measured along a geodesic (i.e.,  shortest
path between the points) as follows:
\begin{equation}\label{d_c(x,y)}
d_{c}(\mathbf{x}, \mathbf{y})=\frac{2}{\sqrt{c}} \tanh ^{-1}\left(\sqrt{c}\left\|-\mathbf{x} \oplus_{c} \mathbf{y}\right\|\right).
\end{equation}

The M$\ddot{o}$bius
substraction is then defined by the use of the following notation: $\bm{x}\ominus_c\bm{y}=\bm{x}\oplus_c(-\bm{y})$.
Similar to the scalar multiplication and matrix multiplication in Euclidean space, the multiplication in hyperbolic space can be defined by M$\ddot{o}$bius scalar multiplication and M$\ddot{o}$bius matrix multiplication between vectors
$\bm{x}\in \mathbb{H}_c^n\{\bm{0}\}$:
\begin{equation}
\begin{aligned}
\bm{r} \otimes_{c} \boldsymbol{x}&=\frac{1}{\sqrt{c}} \tanh \left(r \tanh ^{-1}(\sqrt{c}\|\boldsymbol{x}\|)\right) \frac{\boldsymbol{x}}{\|\boldsymbol{x}\|},\\
\bm{M} \otimes_{c} \boldsymbol{x}&=(\frac{1}{\sqrt{c}})   \tanh \left(\frac{\|\bm{M} \boldsymbol{x}\|}{\|\boldsymbol{x}\|} \tanh ^{-1}(\sqrt{c}\|\boldsymbol{x}\|)\right) \frac{\bm{M} \boldsymbol{x}}{\|\bm{M} \boldsymbol{x}\|},
\end{aligned}
\end{equation}
where $\bm{r}\in \mathbb{R}$ and $\bm{M}\in \mathbb{R}^{m\times n}$.

\section{Methodology}

\subsection{\large{Geometry Interaction Knowledge Graph Embeddings (GIE)}}

\paragraph{Problem Definition.}  Given  a knowledge graph $\mathcal{G}$, 
   practically, it is usually incomplete, then the link prediction approaches are used to predict new links and complete the knowledge graph. Specifically, we learn the embeddings of relations and entities through the existing links among the entities, and then   predict new links based on these embeddings.  Therefore, our ultimate goal is to develop an effective KGE method for the link prediction task.

\paragraph{Representations for Entities and Relations.} In the KGE problem, entities can be  represented as vectors in low-dimensional space while relations can be represented as geometric transformations. In order to portray spatial transformations accurately and comprehensively,   we define the \emph{special orthogonal group} and the \emph{special Euclidean group} as follows respectively:
\begin{definition}
The special orthogonal group is defined as
\begin{equation}\label{special orthogonal group}
\mathbf{S O}(n)=\left\{\mathbf{A} \in \mathbf{G} \mathbf{L}(n, \mathbb{R}) \mid \mathbf{A}^{T} \mathbf{A}=\mathbf{I} \wedge \operatorname{det} \mathbf{A}=1\right\},
\end{equation}
where $\textbf{GL}(n,R)$ denotes the  $n$ dimensional linear group over the set of real numbers $\mathbb{R}$. 
\end{definition}
\begin{definition}
 The special Euclidean group is defined as
 \begin{equation}\label{special Euclidean group}
\mathbf{S E}(n)=\left\{\mathbf{A} \mid \mathbf{A}=\left[\begin{array}{cc}
\mathbf{R} & \mathbf{v} \\
0 & 1
\end{array}\right], \mathbf{R} \in \mathbf{S O}(n), \mathbf{v} \in \mathbb{R}^{n}\right\}.
\end{equation}
Here $\mathbf{SE}(n)$ is the set of all rigid transformations in n dimensions. It provides more degrees of freedom for geometric transformation  than $\mathbf{S O}(n)$.
\end{definition}

Note that the \emph{special Euclidean group} has powerful spatial modeling capabilities and it will play a major role in modeling key inference patterns.  In a geometric perspective,
it supports
multiple transformations, specifically,
reflection, inversion, translation, rotation, and homothety.  We provide the proof in Appendix\footnote{https://github.com/Lion-ZS/GIE}.

\noindent\textbf{Geometry Interaction.} The geometry interaction plays an important role in capturing the reliable structure of the space and we will explain its mechanism below. First of all, geometric message propagation is a prerequisite for geometry interaction. In view of the different spatial properties of Euclidean space, hyperbolic space and hypersphere space,
their spatial structures are established through different spatial metrics (e.g., Euclidean metric or hyperbolic metric), which will affect the geometry
information propagation between these spaces.
Fortunately, the \emph{exponential} and \emph{logarithmic} mappings can be utilized as
bridges between these spaces for geometry
information propagation.

 Then the geometry interaction can be summarized into three steps: (1) We first generate  spatial  embeddings for the entity $\bm{e}$ in the knowledge graph, including an embedding $\bm{E}$ in Euclidean space, an embedding $\bm{H}$ in hyperbolic space and an embeddings $\bm{S}$ in hypersphere space. (2) We map the embeddings $\bm{H}$ and $\bm{S}$ from the hyperbolic  space and hyperspherical space to the tangent space  through \emph{logarithmic} map respectively to facilitate geometry
information propagation. Then we use the attention mechanism to interact the geometric messages of Euclidean space and tangent space. (3)
  We capture the features of different spaces and then adaptively form the reliable spatial structure according to the interactive geometric message.

 Here we discuss an important detail in geometry interaction. Suppose relation  $\bm{r} \in \mathbf{SE}(n)$ and $\bm{r}'$ is the reverse relation of $\bm{r}$\footnote{Suppose $\bm{r}=\left[\begin{array}{cc}
\mathbf{R} & \mathbf{v} \\
0 & 1
\end{array}\right]$, then we can derive $\bm{r'}=\left[\begin{array}{cc}
\mathbf{R}^{-1} & -\mathbf{R}^{-1}\mathbf{v} \\
0 & 1
\end{array}\right]=\left[\begin{array}{cc}
\mathbf{R}^{T} & -\mathbf{R}^{T}\mathbf{v} \\
0 & 1
\end{array}\right]$, which is convenient for calculation  since we only need to calculate the transpose of the matrix instead of the inverse.}.
Notice that the basic operations such as M$\ddot{o}$bius addition does not satisfy commutativity   in hyperbolic space. Therefore, given two points $\bm{x}$ and $\bm{y}$, note that $\bm{x}\oplus_c\bm{y}= \bm{y}\oplus_c\bm{x}$ does not hold for general cases in the hyperbolic space, 
then we have  $\bm{x}\ominus_c \bm{y} \neq -(\bm{y}\ominus_c \bm{x})$. For the triplet $(\bm{h},\bm{r},\bm{t})$ ($\bm{h}$ and $\bm{t}$ denote head entity and tail entity respectively) in KG of non-Euclidean structure, we have:
\begin{equation}
\bm{h}\otimes_c\bm{r}\ominus_c\bm{t}\neq-(\bm{t}\otimes_c\bm{r}'\ominus_c\bm{h}).
\end{equation}

It means that the propagated messages of two opposite directions (i.e., from head entity to tail entity and from tail entity to head entity) are different.
To address it,  we  leverage the embeddings of
head entities and tail entities in different sapces
separately to propagate geometric messages for geometry interaction.  

For the head entity, 
 we have $\bm{E_h}=\bm{rh}$ which  is the embedding of transformed head entity  in  Euclidean space.
At the same time, we have
 $\bm{H_h}=\bm{r}\otimes_v\exp^v_{\bm{0}}(\bm{h})$, which is the embedding of transformed head entity in   hyperbolic space, and $\exp^v_{\bm{0}}(\bm{h})\ (v<0)$ aims to obtain the embedding of the head entity $\bm{h}$ in hyperbolic space. Meanwhile, we have $\bm{S_h}=\bm{r}\otimes_u\exp^u_{\bm{0}}(\bm{h})$, which is the embedding of transformed head entity in   hyperspherical space, and $\exp^u_{\bm{0}}(\bm{h})\ (u>0)$ aims to obtain the embedding of the head entity $\bm{h}$ in the hyperspherical space.
 \begin{equation}\label{inter(h)}
 \begin{aligned}
   \operatorname{Inter}(\bm{E_h},\bm{H_h},\bm{S_h})=\exp_{\bm{0}}^c(\lambda_E\bm{E_h}+\lambda_H\log_{\bm{0}}^v(\bm{H_h})
   \\+\lambda_S\log_{\bm{0}}^u(\bm{S_h})),
 \end{aligned}
 \end{equation}
 where  $\bm{\lambda}$ is an
attention vector and  $\left(\lambda_{E}, \lambda_{H},\lambda_{S}\right)=\operatorname{Softmax}\left(\bm{\lambda}^{T} \bm{E_{h}}, \bm{\lambda}^{T} \bm{H_{h}},\bm{\lambda}^{T} \bm{S_{h}}\right)$.  Here we apply $\log_{\bm{0}}^v(\cdot)$ and $\log_{\bm{0}}^u(\cdot)$  in order to map  $\bm{H}_h$ and $\bm{S_h}$ into the tangent space respectively to propagate geometry information. Moreover, applying $\exp_{\bm{0}}^c(\cdot)$ aims to form an approximate spatial structure for the interactive space.
\begin{table*}[th]
\begin{center}
\begin{tabular}{lccccccc}
  \toprule
   Inference pattern  &TransE &RotatE& TuckER &DistMult &ComplEx& MuRP&GIE \\
     \hline
   Symmetry: $r_1(x,y)\Rightarrow r_1(y,x)$  & $\times$ & $\checkmark$&$\checkmark$&$\checkmark$ &$\checkmark$ &$\times$&$\checkmark$ \\
   Anti-symmery: $r_1(x,y)\Rightarrow \neg r_1(y,x)$& $\checkmark$ &$\checkmark$ &$\checkmark$ &$\times$& $\checkmark$ &$\checkmark$ &$\checkmark$\\
   Inversion: $r_1(x,y) \Leftrightarrow r_2(y,x)$& $\checkmark$ &$\checkmark$ &$\checkmark$ &$\times$& $\checkmark$&$\checkmark$&$\checkmark$\\
   Composition: $r_1(x,y)\wedge r_2(y,z)\Rightarrow r_3(x,z)$ & $\checkmark$ &$\checkmark$ &$\times$ &$\times$ &$\times$&$\checkmark$&$\checkmark$\\
   Hierarchy: $r_1(x,y)\Rightarrow r_2(x,y)$ & $\times$ &$\times$ &$\checkmark$ &$\checkmark$ &$\checkmark$&$\checkmark$&$\checkmark$\\
    Intersection: $r_1(x,y)\wedge r_2(x,y)\Rightarrow r_3(x,y)$ &$\checkmark$ &$\checkmark$&$\times$ &$\times$& $\times$&$\checkmark$&$\checkmark$\\
    Mutual exclusion: $r_1(x,y)\wedge r_2(x,y)\Rightarrow \perp$ &$\checkmark$ &$\checkmark$ &$\checkmark$ &$\checkmark$&$\checkmark$  &$\checkmark$&$\checkmark$\\
    \bottomrule
\end{tabular}
\caption{\label{Inference patterns} Inference patterns captured by some popular KGE models.}
\end{center}
\end{table*}

 To preserve more information of Euclidean, hyperspherical and hyperbolic spaces, we aggregate their messages through performing self-attention guided by geometric distance between entities. After
aggregating messages, we apply a pointwise non-linearity function to measure their messages.  Then, we  form a suitable spatial structure based on the geometric message. In this way,  we have the transformed entity interacting in Euclidean space, hyperspherical space and hyperbolic space as follows:

Similarly, for the tail entity, 
let $\bm{E_t}=\bm{r't}$, $\bm{H_t}=\bm{r}'\otimes_v\exp^v_{\bm{0}}(\bm{t})$ and $\bm{S_t}=\bm{r}'\otimes_u\exp^u_{\bm{0}}(\bm{t})$ be the transformed tail entities in  Euclidean space, hyperbolic space and hyperspherical space, respectively. Then we  aggregate their messages  interacting in the space  as follows:
 \begin{equation}\label{inter(t)}
 \begin{aligned}
   \operatorname{Inter}(\bm{E_t},\bm{H_t},\bm{S_t})=\exp_{\bm{0}}^c(\alpha_E\bm{E_t}+\alpha_H\log_{\bm{0}}^v(\bm{H_t})
   \\+\alpha_S\log_{\bm{0}}^u(\bm{S_t})),
 \end{aligned}
 \end{equation}
 where  $\bm{\alpha}$ is an
attention vector and  $\left(\alpha_{E}, \alpha_{H},\alpha_{S}\right)=\operatorname{Softmax}\left(\bm{\alpha}^{T} \bm{E_{t}}, \bm{\alpha}^{T} \bm{H_{t}},\bm{\alpha}^{T} \bm{S_{t}}\right)$.

\paragraph{Score Function and Loss Function.}  
After integrating the information for head entities and tail entities in different spaces respectively,  
 we design the scoring function for geometry interaction as follows:
\begin{equation}\label{score function}
\begin{aligned}
\phi(\bm{h},\bm{r},\bm{t})=&-(d_{c}(Inter(\bm{E_h},\bm{H_h},\bm{S_h}), \bm{t})\\&+d_c(Inter(\bm{E_t},\bm{H_t},\bm{S_t}),\bm{h}))+\bm{b}.
\end{aligned}
\end{equation}
where $d_c(\cdot,\cdot)$  represents the distance function, 
 and $\bm{b}$ is the bias which act as the margin in the scoring function \cite{2019Multi}.

 Here we train GIE by minimizing the cross-entropy
loss with uniform negative sampling, where negative examples for a triplet $(\bm{h},\bm{r},\bm{t})$ are sampled
uniformly from all possible triplets obtained by perturbing the tail entity:
\begin{equation}\mathcal{L}=\sum_{\{\bm{h},\bm{r},\bm{t}\} \in \Omega \cup \Omega'} \log \left(1+\exp \left(-Y\phi(\bm{h},\bm{r},\bm{t}) \right)\right),\end{equation}
where $\Omega$ and $\Omega'=\mathcal{E}\times \mathcal{R}\times \mathcal{E}-\Omega$  denote the set of
observed triplets and the set of unobserved triplets, respectively. $Y\in \{-1,1\}$ represents the corresponding label of the triplet $(\bm{h},\bm{r},\bm{t})$.

\subsection{Key Properties of GIE}

\textbf{Connection to  Euclidean space, hyperbolic space and hyperspherical space.} The geometry interaction mechanism allows us to flexibly take advantages of Euclidean space, hyperbolic space and hyperspherical space. For the chain structure in KGs, we can adjust the local structure of the embedding space in geometry interaction to be closer to Euclidean space.
Recall the two maps in Eq.(\ref{two maps}) have more appealing forms when $\bm{x} = \bm{0}$, namely for $ \bm{v} \in T_{\mathbf{0}} \mathbb{D}_{c}^{n} \backslash\{\mathbf{0}\}, \bm{y} \in \mathbb{D}_{c}^{n} \backslash\{\mathbf{0}\}$:
$\exp _{\mathbf{0}}^{c}(\bm{v})=\tanh (\sqrt{c}\|\bm{v}\|) \frac{\bm{v}}{\sqrt{c}\|\bm{v}\|}, \log _{\boldsymbol{0}}^{c}(\bm{y})=\tanh ^{-1}(\sqrt{c}\|\bm{y}\|) \frac{\bm{y}}{\sqrt{c}\|\bm{y}\|}.
$ 
 Moreover, we can recover Euclidean geometry 
as $\lim_{ c\rightarrow0} \exp^c_{\bm{x}} (\bm{v}) = \bm{x} + \bm{v}$, 
 and $\lim_{c\rightarrow0}\log^c_{\bm{x}}(\bm{y}) = \bm{y} - \bm{x}$.  For the hierarchical structure, we can make the local structure of the embedded space close to the hyperbolic structure through geometry interaction (make the space curvature $c<0$).
 For the ring structure, we can form the hyperspherical space ($c>0$) and the process is similar to the above.

If the embedding locations of entities have multiple structural features  (e.g., hierarchy structure and ring  structure) simultaneously,
we  can integrate different geometric messages as shown in the Eq.(\ref{inter(h)}) and Eq.(\ref{inter(t)}), then
accordingly form the reliable spatial structure by geometry interaction  as above.

\paragraph{\textbf{Capability in Modeling Inference Patterns.}} Generally,  knowledge graphs embrace multiple relations, which  forms the basis of pattern reasoning. Therefore  it is particularly important to enable inference  patterns applied in different spaces.
Suppose $\bm{M}$ is a relation matrix, and then
take the  entity $\bm{e}$ as an example. If we have
$f_k(\bm{e})=\phi_k(\bm{M}_k\bm{e})$, it comes the M$\ddot{o}$bius version  $f_{k}^{\otimes_{c}}(\bm{e})=\varphi_{k}^{\otimes_{c}}\left(\bm{M}_{k} \otimes_{c} \bm{e}\right)$, where $f_k$ can be any function  mapping we need. Then their composition can be rewritten as
$f_{k}^{\otimes_{c}} \circ \cdots \circ f_{1}^{\otimes_{c}}=\exp _{\bm{0}}^{c} \circ f_{k} \circ \cdots \circ f_{1} \circ \log _{\bm{0}}^{c},
$ 
which means these operations in other spaces can be essentially
performed in Euclidean space. 

 Similar to  other M$\ddot{o}$bius operations,
 for a given function in non-Euclidean space, we can restore the form of the function in Euclidean space
 by limiting $c\rightarrow0$,  which can be formulated as $\lim_{c\rightarrow0}f^{\otimes_c}(\bm{x})=f(\bm{x})$ if $f$ is  continuous. This definition can also satisfy some desirable properties, such as $(f\circ g)^{\otimes_c}=f^{\otimes_c}\circ g^{\otimes_c}$ for $f:\mathbb{R}\rightarrow \mathbb{R}^l$ and $g:\mathbb{R}^n\rightarrow \mathbb{R}^m$, 
 and $f^{\otimes_{c}}(\bm{x}) /\left\|f^{\otimes_{c}}(\bm{x})\right\|=f(\bm{x}) /\|f(\bm{x})\|$ for $f(\bm{x}) \neq \mathbf{0}$ (direction preserving).

 \begin{table*}[t]
\begin{center}
 \begin{tabular}{ccccccccc}
  \toprule
Dataset & N & M & Training & Validation & Test &$\xi_G$ & Structure Heterogeneity& Scale\\
  \midrule
 WN18RR & 41k& 11& 87k& 3k& 3k& -2.54& Low & Small\\
FB15k-237 &15k &237& 272k& 18k &20k& -0.65&High & Medium\\
YAGO3-10 &  123k &37& 1M& 5k& 5k&-0.54& Low& Large\\
  \bottomrule
 \end{tabular}
 \caption{\label{data}Statistics of the datasets used in this paper. (N is the number of entities and M is the number of relations. Refer to Appendix \ref{Hierarchy estimates} for more details about $\xi_G$.)
}
 \end{center}
\end{table*}

Based on the analysis above, 
we can extend the logical rules established in Euclidean space  to the hyperbolic space and hyperspheric space in geometry interaction. In view of this, 
geometry interaction enables us to model key relation patterns in different spaces conveniently. As shown in Table \ref{Inference patterns}, there is a comparison of GIE against these models
with respect to capturing prominent inference patterns, which shows the advantages of  GIE in modeling these patterns. 
Please refer to  Appendix  for the proof.

\section{Experiments}
\subsection{Experimental Setup}
 \paragraph{Dataset.} We evaluate our approach on the link prediction task using three standard competition benchmarks as shown in Table \ref{data}, namely  WN18RR \cite{dettmers2017convolutional}, FB15K-237 \cite{dettmers2017convolutional} and YAGO3-10 \cite{2013YAGO3}. WN18RR is a subset of WN18 \cite{bordes2013translating}  and the main relation patterns are symmetry/antisymmetry and composition.
FB15K237  is a subset of FB15K \cite{bordes2013translating}, in which the inverse relations  are removed.
YAGO3-10, a subset of YAGO3 \cite{DBLP:conf/cidr/MahdisoltaniBS15}, mainly includes  symmetry/antisymmetry and composition patterns. Each dataset is split into training,
validation and testing sets, which is the same as the setting of \cite{sun2019rotate:}. For each KG, we follow the standard data augmentation protocol by adding inverse relations
\cite{lacroix2018canonical} to the datasets for baselines.

 \paragraph{Evaluation metrics.}
For each case, the test triplets are ranked among all triplets
with masked entities replaced by entities in the knowledge graph. 
Hit@n (n=1,3,10) and the Mean Reciprocal
Rank (MRR) are reported in the experiments. We follow the standard evaluation
protocol in the filtered setting \cite{bordes2013translating}:
all true triplets in the KG are filtered out during
evaluation, since predicting a low rank for these
triplets should not be penalized.

 \paragraph{Baselines.} We compare our method to SOTA models, including Euclidean approaches based on TransE \cite{bordes2013translating} and DistMult \cite{yang2014embedding};
  complex approaches based on
TuckER \cite{balavzevic2019tucker},
ComplEx \cite{trouillon2016complex},
and RotatE \cite{sun2019rotate:}; 
 Non-Euclidean  approaches based on  MuRP \cite{2019Multi} and    \textsc{AttH} \cite{chami2020low}; 
  and  neural network approaches based on ConvE \cite{dettmers2017convolutional},
R-GCN \cite{schlichtkrull2018modeling} and  HypER \cite{balavzevic2019hypernetwork}.

\paragraph{Implementation Details.} We implement GIE in PyTorch and run experiments on a single GPU.
The hyper-parameters are determined by the grid search. The best models are selected by early stopping
on the validation set. In general, the embedding size $k$ is searched in $\{50,100,200,250\}$.
 Learning rate is  tuned amongst $\{0.001,0.005,0.05,0.1\}$.
 For some baselines, we report the results in the original papers. 

\begin{table*}
\begin{center}
 \begin{tabular}{c|c|cccc|cccc|cccc}
  \toprule
 & $\ $ &  \multicolumn{4}{c}{WN18RR}  &\multicolumn{4}{c}{FB15k-237}&\multicolumn{4}{c}{YAGO3-10} \\
    $\mathcal{M}$&Model  &MRR &H@1 &H@3 &H@10  &MRR &H@1 &H@3 &H@10&MRR &H@1 &H@3 &H@10\\
  \midrule
 $\mathbb{R}$& RESCAL &.420 &- &-& .447& .270& - &- &-& -& - &- &-\\
$\mathbb{R}$& TransE &.226 &- &-& .501& .294& - &- &.465& -& - &- &-\\
$\mathbb{R}$& DisMult &.430 &.390 &.440 &.490 &.241 &.155 &.263 &.419 &.340 &.240 &.380 &.540\\
$\mathbb{R}$& ConvE &.430 &.400 &.440 &.520 &.325 &.237& .356& .501& .440& .350& .490& .620\\
$\mathbb{R}$&TuckER& .470& .443& .482& .526& .\underline{358}& \underline{.266}& \underline{.394}& \underline{.544}& -& -& -& -\\
$\mathbb{R}$& MuRE &.465 &.436 &.487 &.554 &.336 &.245 &.370 &.521 &.532 &.444 &.584 &.694\\
$\mathbb{R}$& CompGCN &.479 &.443 &.494 &.546 &.355 &.264 &.390 &.535 &.489 &.395 &.500 &.582\\
$\mathbb{R}$& A2N &.430 &.410 &.440 &.510 &.317 &.232 &.348 &.486 &.445 &.349 &.482 &.501\\
$\mathbb{C}$& ComplEx &.440 &.410 &.460 &.510 &.247 &.158 &.275 &.428 &.360 &.260 &.400 &.550\\
$\mathbb{C}$ &RotatE &.476 &.428 &.492 &.571 &.338 &.241 &.375 &.533 &.495 &.402 &.550 &.670\\
$\mathbb{H}$ &MuRP &.481 &.440 &.495 &.566 &.335 &.243 &.367 &.518 &.354 &.249 &.400 &.567\\
$\mathbb{H}$ &ATTH &\underline{.486} &\underline{.443} &\underline{.499} &\underline{.573} &.348 &.252 &.384 &.540 &\underline{.568}&\underline{.493} &\underline{.612} &\underline{.702} \\
$\mathbb{S}$ &MuRS &.454 &.432 &.482 &.550 &.338 &.249 &.373 &.525 &.351 &.244 &.382 &.562\\
$\mathbb{P}$ &MuRMP &.473 &.435 &.485 &.552 &.345 &.258 &.385 &.542 &.358 &.248 &.389 &.566\\
    \midrule
$\mathbb{G}$&\textbf{GIE}    
&$\bm{.491}$   &   $\bm{.452}$&$\bm{.505}$  &$\bm{.575}$
&$\bm{.362}$&$\bm{.271}$&$\bm{.401}$&$\bm{.552}$  & $\bm{.579}$ & $\bm{.505}$  &$\bm{.618}$ &$\bm{.709}$\\

  \bottomrule
 \end{tabular}
 \caption{ Link prediction results on three datasets. $\mathbb{R}$ represents the Euclidean space. $\mathbb{C}$ represents the complex space. 
$\mathbb{H}$ represents the hyperbolic space. 
 $\mathbb{S}$ represents the spherical space. $\mathbb{P}$ represents the mix-curvature space. $\mathbb{G}$ represents the geometric interactive space. Best results are in bold and second best
results are underlined.}\label{WN18RR}
 \end{center}
\end{table*}

\subsection{Results}

 We show empirical results of three datasets in Table \ref{WN18RR}.
On WN18RR  where there are plenty of symmetry and composition relations, MRR is improved from $48.1\%$ (MuRP) to $49.1\%$, which reflects the effective modeling capability of GIE for key patterns. WN18RR contains rich hierarchical data, so it can be seen that GIE can model the hyperbolic structure well.
On FB15k237,  GIE
 outperforms on MRR and yields the score of
$36.2\%$ while previous work based on the Non-Euclidean space  (MuRP) can only reach $33.5\%$ at most.
 On YAGO3-10, which  has  about 1,000,000 training samples,
  MRR is improved from $56.8\%$ to $57.9\%$. This shows that   GIE can learn embedding well for the  knowledge graphs with large scale.

It can be viewed that GIE  performs best compared to the existing state-of-the-art models across  on all datasets above. The reason lies in that the structures in KGs  are usually complex, which pose a challenge for learning  embedding. In this case, GIE can learn a more reliable structure with geometry interaction while the structure captured by the previous work are relatively inaccurate and incomplete.

\subsection{Experiment Analysis}

\noindent\textbf{Ablations on Relation Type.} In Table \ref{11 relations} in Appendix, we summarize the Hits@10 for each relation on WN18RR to confirm
the superior representation capability of GIE in modelling different types of relation. 
 We show how the relation types on WN18RR affect the performance of the proposed method and some baselines such as MuRMP-autoKT, MuRMP, MuRS, MuRP and MuRE. We report a number of metrics to describe each relation, including global graph curvature ($\xi_G$) \cite{DBLP:conf/iclr/GuSGR19} and Krackhardt hierarchy score
(Khs) \cite{DBLP:journals/socnet/EverettK12}, which can justify whether the given data have a rich hierarchical structure. Specifically, we compare averaged hits@10 over 10 runs for models
  of low dimensionality. 
 We can see that the proposed  model performs well for different types of relations, which verifies the effectiveness of the proposed method in dealing with complex structures in KGs.
Moreover, we can see that the geometry interaction is more effective than the  mixed-curvature geometry method. 

\begin{figure}
	\centering
\includegraphics[width=2.4in]{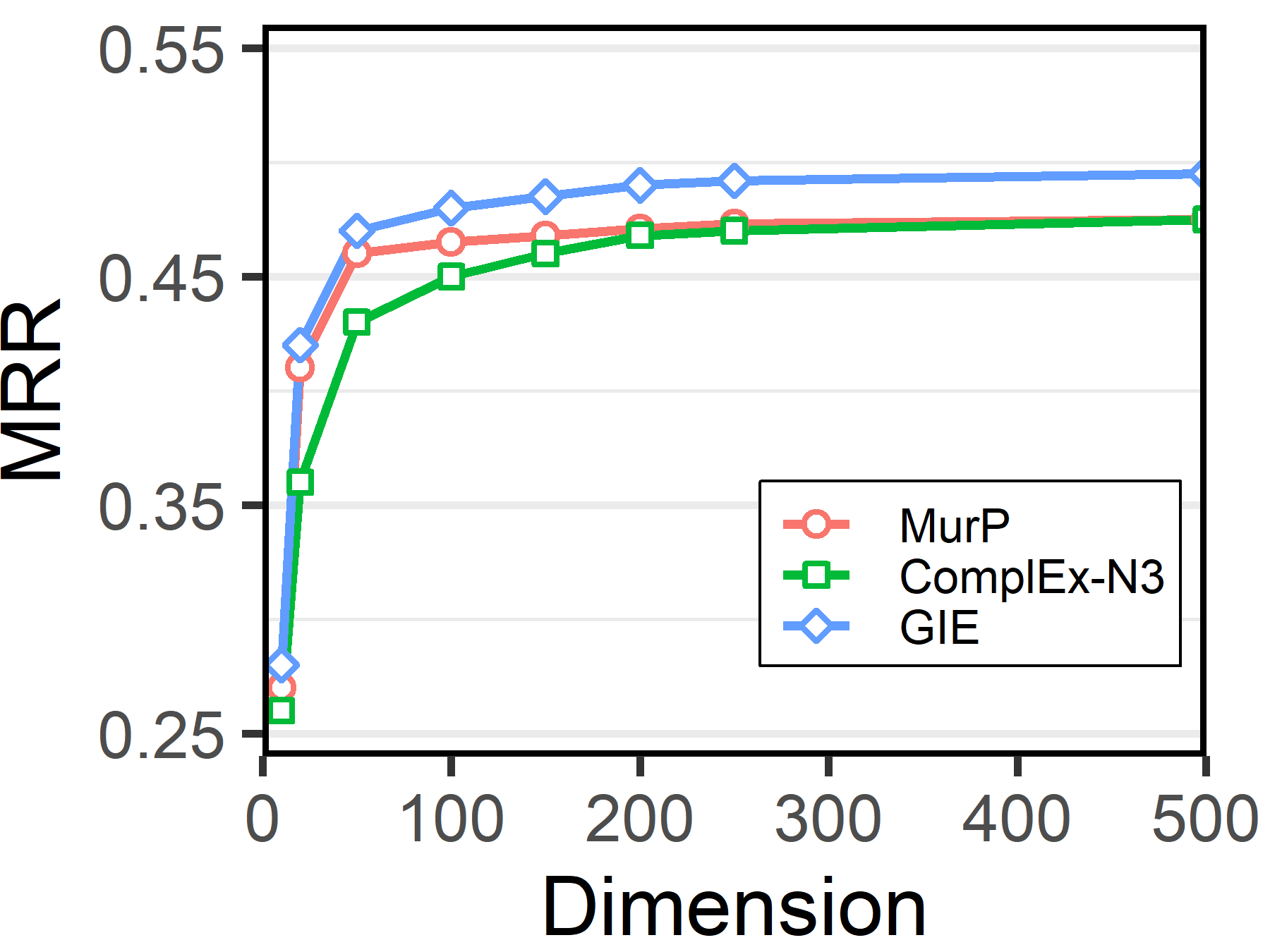}
\caption{ Impact of embedding dimension.}\label{Impact of embedding dimension}
\end{figure}
\paragraph{Impact of Embedding Dimension.} To understand the role of dimensionality, we conduct experiments on WN18RR against
SOTA methods under varied dimensional settings $(10,20,50,100,150,200,250,500)$ shown in Figure \ref{Impact of embedding dimension}. GIE can  exploit hyperbolic, hyperspheric
and Euclidean spaces jointly for knowledge graph embedding through an interaction learning
mechanism, and we  can see that GIE  performs well in both low and high dimensions.

\begin{table}
\begin{center}
 \begin{tabular}{lcccc}
  \toprule
  Model & RotatE & MuRP  &GIE\\
  \midrule
WN18RR  & 40.95M &32.76M &17.65M ($\downarrow 46.1\%$)\\
FB15K237 &29.32M &5.82M&3.41M ($\downarrow 41.4\%$)\\
YAGO3-10& 49.28M &41.47M &23.81M ($\downarrow 42.6\%$)\\
  \bottomrule
 \end{tabular}
 \end{center}
 \caption{Number of free parameters comparison.
  }
  \label{Number of free parameters comparion}
\end{table}
\paragraph{Number of Free Parameters Comparison.} We compare the numbers of parameters of  RotatE and MuRP with that of GIE in the experiments above. Geometry interaction can
  learn a reliable spatial structure by geometric interaction, which reduce the dependence on adding dimensions to improve performance. 
 As shown in Table \ref{Number of free parameters comparion}, it obviously shows that GIE can maintain superior performance with fewer parameters.

\noindent\textbf{Ablation Study on Geometry Interaction.} In order to study the performance of each component  in the geometry interaction score function Eq.(\ref{score function}) separately,  we  remove $d_c(\operatorname{Inter}(\bm{E_t},\bm{H_t},\bm{S_t}),\bm{h})$ of the score function and denote the model as GIE$_1$. We  remove $d_c(\operatorname{Inter}(\bm{E_h},\bm{H_h},\bm{S_h}),\bm{t})$  of the score function and denote the model as GIE$_2$. We remove the geometry interaction process of the score function and denote the model as GIE$_3$.
As shown in Table \ref{Ablation Study on Score Function} in Appendix,
GIE$_1$, GIE$_2$ and GIE$_3$ perform poorly compared to GIE. The experiment proves
that
the geometry interaction is a critical step for the embedding performance since geometric interaction is  extremely beneficial for learning reliable spatial structures of KGs.

\section{Conclusion}
In this paper, we design a new knowledge graph embedding model named GIE that can  exploit structures of knowledge
graphs with interaction in Euclidean, hyperbolic and hyperspherical space simultaneously. Theoretically , GIE is advantageous with its
capability in learning
reliable spatial structure features in an adaptive way.
Moreover, GIE can  embrace rich and
expressive semantic matching between entities and satisfy the key of relational representation learning.
 Empirical experimental evaluations on three well-established
datasets show that GIE can achieve an overall the state-of-the-art performance, outperforming multiple
recent strong baselines with fewer free parameters.

\newpage

\section{Acknowledgements}
This work was supported in part by the National Key R$\&$D Program of China under Grant 2018AAA0102003, in part by National Natural Science Foundation of China: U21B2038, U1936208, 61931008, 61836002, 61733007, 6212200758 and 61976202,
in part by the Fundamental Research Funds for the Central Universities,
 in part by the National Postdoctoral Program for Innovative Talents, Grant No. BX2021298,
 in part by Youth Innovation Promotion Association CAS, and in part by the Strategic Priority Research Program of Chinese Academy of Sciences, Grant No. XDB28000000.

\bibliography{GIE} 

\begin{thebibliography}{45}
\providecommand{\natexlab}[1]{#1}

\bibitem[{Balazevic, Allen, and
  Hospedales(2019{\natexlab{a}})}]{balavzevic2019hypernetwork}
Balazevic, I.; Allen, C.; and Hospedales, T.~M. 2019{\natexlab{a}}.
\newblock Hypernetwork Knowledge Graph Embeddings.
\newblock In \emph{ICANN}, 553--565.

\bibitem[{Balazevic, Allen, and Hospedales(2019{\natexlab{b}})}]{2019Multi}
Balazevic, I.; Allen, C.; and Hospedales, T.~M. 2019{\natexlab{b}}.
\newblock Multi-relational Poincar{\'{e}} Graph Embeddings.
\newblock In \emph{NeurIPS}, 4465--4475.

\bibitem[{Bala{\v{z}}evi{\'c}, Allen, and
  Hospedales(2019)}]{balavzevic2019tucker}
Bala{\v{z}}evi{\'c}, I.; Allen, C.; and Hospedales, T.~M. 2019.
\newblock Tucker: Tensor factorization for knowledge graph completion.
\newblock In \emph{EMNLP}, 5185--5194.

\bibitem[{Bansal et~al.(2019)Bansal, Juan, Ravi, and McCallum}]{bansal2019a2n}
Bansal, T.; Juan, D.-C.; Ravi, S.; and McCallum, A. 2019.
\newblock A2N: attending to neighbors for knowledge graph inference.
\newblock In \emph{ACL}, 4387--4392.

\bibitem[{Berant et~al.(2013)Berant, Chou, Frostig, and
  Liang}]{DBLP:conf/emnlp/BerantCFL13}
Berant, J.; Chou, A.; Frostig, R.; and Liang, P. 2013.
\newblock Semantic Parsing on Freebase from Question-Answer Pairs.
\newblock In \emph{ACL}, 1533--1544.

\bibitem[{Berant and Liang(2014)}]{DBLP:conf/acl/BerantL14}
Berant, J.; and Liang, P. 2014.
\newblock Semantic Parsing via Paraphrasing.
\newblock In \emph{ACL}, 1415--1425.

\bibitem[{Bordes et~al.(2013)Bordes, Usunier, Garcia-Duran, Weston, and
  Yakhnenko}]{bordes2013translating}
Bordes, A.; Usunier, N.; Garcia-Duran, A.; Weston, J.; and Yakhnenko, O. 2013.
\newblock Translating embeddings for modeling multi-relational data.
\newblock In \emph{NeurIPS}, 2787--2795.

\bibitem[{Cao et~al.(2021)Cao, Xu, Yang, Cao, and
  Huang}]{DBLP:conf/aaai/CaoX0CH21}
Cao, Z.; Xu, Q.; Yang, Z.; Cao, X.; and Huang, Q. 2021.
\newblock Dual Quaternion Knowledge Graph Embeddings.
\newblock In \emph{AAAI}, 6894--6902.

\bibitem[{Chami et~al.(2020)Chami, Wolf, Juan, Sala, Ravi, and
  R{\'e}}]{chami2020low}
Chami, I.; Wolf, A.; Juan, D.-C.; Sala, F.; Ravi, S.; and R{\'e}, C. 2020.
\newblock Low-Dimensional Hyperbolic Knowledge Graph Embeddings.
\newblock In \emph{ACL}.

\bibitem[{Dettmers et~al.(2017)Dettmers, Minervini, Stenetorp, and
  Riedel}]{dettmers2017convolutional}
Dettmers, T.; Minervini, P.; Stenetorp, P.; and Riedel, S. 2017.
\newblock Convolutional 2D Knowledge Graph Embeddings.
\newblock In \emph{AAAI}, 1811--1818.

\bibitem[{Diefenbach, Singh, and Maret(2018)}]{DBLP:conf/www/DiefenbachSM18}
Diefenbach, D.; Singh, K.~D.; and Maret, P. 2018.
\newblock WDAqua-core1: {A} Question Answering service for {RDF} Knowledge
  Bases.
\newblock In \emph{WWW}, 1087--1091.

\bibitem[{Dong et~al.(2014)Dong, Gabrilovich, Heitz, Horn, and
  Zhang}]{2014Knowledge}
Dong, X.; Gabrilovich, E.; Heitz, G.; Horn, W.; and Zhang, W. 2014.
\newblock \emph{Knowledge vault: A web-scale approach to probabilistic
  knowledge fusion}.
\newblock ACM.

\bibitem[{Ebisu and Ichise(2018)}]{ebisu2018toruse:}
Ebisu, T.; and Ichise, R. 2018.
\newblock TorusE: Knowledge Graph Embedding on a Lie Group.
\newblock In \emph{AAAI}, 1819--1826.

\bibitem[{Everett and Krackhardt(2012)}]{DBLP:journals/socnet/EverettK12}
Everett, M.~G.; and Krackhardt, D. 2012.
\newblock A second look at Krackhardt's graph theoretical dimensions of
  informal organizations.
\newblock \emph{Soc. Networks}, 34(2): 159--163.

\bibitem[{Ganea, B{\'{e}}cigneul, and Hofmann(2018)}]{DBLP:conf/nips/GaneaBH18}
Ganea, O.; B{\'{e}}cigneul, G.; and Hofmann, T. 2018.
\newblock Hyperbolic Neural Networks.
\newblock In \emph{NeurIPS}, 5350--5360.

\bibitem[{Gong et~al.(2020)Gong, Wang, Wang, Feng, Peng, Tang, and
  Yu}]{DBLP:conf/sigir/Gong0WFP0Y20}
Gong, J.; Wang, S.; Wang, J.; Feng, W.; Peng, H.; Tang, J.; and Yu, P.~S. 2020.
\newblock Attentional Graph Convolutional Networks for Knowledge Concept
  Recommendation in MOOCs in a Heterogeneous View.
\newblock In \emph{SIGIR}, 79--88.

\bibitem[{Gu et~al.(2019{\natexlab{a}})Gu, Sala, Gunel, and
  R{\'{e}}}]{DBLP:conf/iclr/GuSGR19}
Gu, A.; Sala, F.; Gunel, B.; and R{\'{e}}, C. 2019{\natexlab{a}}.
\newblock Learning Mixed-Curvature Representations in Product Spaces.
\newblock In \emph{ICLR}.

\bibitem[{Gu et~al.(2019{\natexlab{b}})Gu, Sala, Gunel, and
  Re}]{LCLR2019Mainfold}
Gu, A.; Sala, F.; Gunel, B.; and Re, C. 2019{\natexlab{b}}.
\newblock LEARNING MIXED-CURVATURE REPRESENTATIONS IN PRODUCTS OF MODEL SPACES.
\newblock In \emph{ICLR}.

\bibitem[{Guo et~al.(2020)Guo, Zhuang, Qin, Zhu, Xie, Xiong, and
  He}]{DBLP:journals/corr/abs-2003-00911}
Guo, Q.; Zhuang, F.; Qin, C.; Zhu, H.; Xie, X.; Xiong, H.; and He, Q. 2020.
\newblock A Survey on Knowledge Graph-Based Recommender Systems.
\newblock \emph{CoRR}, abs/2003.00911.

\bibitem[{He et~al.(2017)He, Balakrishnan, Eric, and
  Liang}]{DBLP:conf/acl/HeBEL17}
He, H.; Balakrishnan, A.; Eric, M.; and Liang, P. 2017.
\newblock Learning Symmetric Collaborative Dialogue Agents with Dynamic
  Knowledge Graph Embeddings.
\newblock In \emph{ACL}, 1766--1776.

\bibitem[{He et~al.(2015)He, Liu, Ji, and Zhao}]{He2015Learning}
He, S.; Liu, K.; Ji, G.; and Zhao, J. 2015.
\newblock Learning to represent knowledge graphs with gaussian embedding.
\newblock In \emph{CIKM}, 623--632.

\bibitem[{Ji et~al.(2015)Ji, He, Xu, Liu, and Zhao}]{ji2015knowledge}
Ji, G.; He, S.; Xu, L.; Liu, K.; and Zhao, J. 2015.
\newblock Knowledge Graph Embedding via Dynamic Mapping Matrix.
\newblock In \emph{ACL}, 687--696.

\bibitem[{Jin et~al.(2021{\natexlab{a}})Jin, Huo, Liang, and
  Yang}]{DBLP:conf/www/JinHL021}
Jin, D.; Huo, C.; Liang, C.; and Yang, L. 2021{\natexlab{a}}.
\newblock Heterogeneous Graph Neural Network via Attribute Completion.
\newblock In \emph{WWW}, 391--400.

\bibitem[{Jin et~al.(2021{\natexlab{b}})Jin, Yu, Jiao, Pan, Yu, and
  Zhang}]{DBLP:journals/corr/abs-2101-01669}
Jin, D.; Yu, Z.; Jiao, P.; Pan, S.; Yu, P.~S.; and Zhang, W.
  2021{\natexlab{b}}.
\newblock A Survey of Community Detection Approaches: From Statistical Modeling
  to Deep Learning.
\newblock \emph{IEEE Transactions on Knowledge and Data Engineering}, DOI:
  10.1109/TKDE.2021.3104155.

\bibitem[{Keizer et~al.(2017)Keizer, Guhe, Cuay{\'{a}}huitl, Efstathiou,
  Engelbrecht, Dobre, Lascarides, and Lemon}]{DBLP:conf/eacl/LascaridesLGKCE17}
Keizer, S.; Guhe, M.; Cuay{\'{a}}huitl, H.; Efstathiou, I.; Engelbrecht, K.;
  Dobre, M.~S.; Lascarides, A.; and Lemon, O. 2017.
\newblock Evaluating Persuasion Strategies and Deep Reinforcement Learning
  methods for Negotiation Dialogue agents.
\newblock In \emph{EACL}, 480--484.

\bibitem[{Lacroix, Usunier, and Obozinski(2018)}]{lacroix2018canonical}
Lacroix, T.; Usunier, N.; and Obozinski, G. 2018.
\newblock Canonical Tensor Decomposition for Knowledge Base Completion.
\newblock In \emph{ICML}, 2863--2872.

\bibitem[{Lin et~al.(2015)Lin, Liu, Sun, Liu, and Zhu}]{lin2015learning}
Lin, Y.; Liu, Z.; Sun, M.; Liu, Y.; and Zhu, X. 2015.
\newblock Learning entity and relation embeddings for knowledge graph
  completion.
\newblock In \emph{AAAI}, 2181--2187.

\bibitem[{Mahdisoltani, Biega, and Suchanek(2013)}]{2013YAGO3}
Mahdisoltani, F.; Biega, J.; and Suchanek, F. 2013.
\newblock YAGO3: A Knowledge Base from Multilingual Wikipedias.
\newblock In \emph{CIDR}.

\bibitem[{Mahdisoltani, Biega, and
  Suchanek(2015)}]{DBLP:conf/cidr/MahdisoltaniBS15}
Mahdisoltani, F.; Biega, J.; and Suchanek, F.~M. 2015.
\newblock {YAGO3:} {A} Knowledge Base from Multilingual Wikipedias.
\newblock In \emph{CIDR}.

\bibitem[{Nathani et~al.(2019)Nathani, Chauhan, Sharma, and
  Kaul}]{nathani2019learning}
Nathani, D.; Chauhan, J.; Sharma, C.; and Kaul, M. 2019.
\newblock Learning Attention-based Embeddings for Relation Prediction in
  Knowledge Graphs.
\newblock In \emph{ACL}, 4710--4723.

\bibitem[{Nguyen et~al.(2018)Nguyen, Nguyen, Nguyen, and
  Phung}]{nguyen2017novel}
Nguyen, D.~Q.; Nguyen, T.~D.; Nguyen, D.~Q.; and Phung, D. 2018.
\newblock A novel embedding model for knowledge base completion based on
  convolutional neural network.
\newblock In \emph{NAACL-HLT}, 327--333.

\bibitem[{Nickel, Rosasco, and Poggio(2016)}]{nickel2015holographic}
Nickel, M.; Rosasco, L.; and Poggio, T. 2016.
\newblock Holographic Embeddings of Knowledge Graphs.
\newblock In \emph{AAAI}, 1955--1961.

\bibitem[{Nickel, Tresp, and Kriegel(2011)}]{2011RESCAL}
Nickel, M.; Tresp, V.; and Kriegel, H.~P. 2011.
\newblock A Three-Way Model for Collective Learning on Multi-Relational Data.
\newblock In \emph{ICML}.

\bibitem[{Schlichtkrull et~al.(2018)Schlichtkrull, Kipf, Bloem, Van Den~Berg,
  Titov, and Welling}]{schlichtkrull2018modeling}
Schlichtkrull, M.; Kipf, T.~N.; Bloem, P.; Van Den~Berg, R.; Titov, I.; and
  Welling, M. 2018.
\newblock Modeling relational data with graph convolutional networks.
\newblock In \emph{ESWC}, 593--607. Springer.

\bibitem[{Socher et~al.(2013)Socher, Chen, Manning, and Ng}]{2013Reasoning}
Socher, R.; Chen, D.; Manning, C.~D.; and Ng, A.~Y. 2013.
\newblock Reasoning With Neural Tensor Networks for Knowledge Base Completion.
\newblock In \emph{NeurIPS}.

\bibitem[{Sun et~al.(2019)Sun, Deng, Nie, and Tang}]{sun2019rotate:}
Sun, Z.; Deng, Z.; Nie, J.; and Tang, J. 2019.
\newblock RotatE: Knowledge Graph Embedding by Relational Rotation in Complex
  Space.
\newblock In \emph{ICLR}, 1--18.

\bibitem[{Tang et~al.(2020)Tang, Huang, Wang, He, and
  Zhou}]{tang-etal-2020-orthogonal}
Tang, Y.; Huang, J.; Wang, G.; He, X.; and Zhou, B. 2020.
\newblock Orthogonal Relation Transforms with Graph Context Modeling for
  Knowledge Graph Embedding.
\newblock In \emph{ACL}, 2713--2722.

\bibitem[{Trouillon et~al.(2016)Trouillon, Welbl, Riedel, Gaussier, and
  Bouchard}]{trouillon2016complex}
Trouillon, T.; Welbl, J.; Riedel, S.; Gaussier, E.; and Bouchard, G. 2016.
\newblock Complex embeddings for simple link prediction.
\newblock In \emph{ICML}, 2071--2080.

\bibitem[{Wang et~al.(2021)Wang, Wei, dos Santos, Wang, Nallapati, Arnold,
  Xiang, Philip, and Cruz}]{wang2021mixed}
Wang, S.; Wei, X.; dos Santos, C.~N.; Wang, Z.; Nallapati, R.; Arnold, A.;
  Xiang, B.; Philip, S.~Y.; and Cruz, I.~F. 2021.
\newblock Mixed-Curvature Multi-Relational Graph Neural Network for Knowledge
  Graph Completion.
\newblock In \emph{WWW}.

\bibitem[{Wang et~al.(2014)Wang, Zhang, Feng, and Chen}]{wang2014knowledge}
Wang, Z.; Zhang, J.; Feng, J.; and Chen, Z. 2014.
\newblock Knowledge graph embedding by translating on hyperplanes.
\newblock In \emph{AAAI}, 1112--1119.

\bibitem[{Xiao, Huang, and Zhu(2016{\natexlab{a}})}]{DBLP:conf/ijcai/0005HZ16}
Xiao, H.; Huang, M.; and Zhu, X. 2016{\natexlab{a}}.
\newblock From One Point to a Manifold: Knowledge Graph Embedding for Precise
  Link Prediction.
\newblock In \emph{IJCAI}, 1315--1321.

\bibitem[{Xiao, Huang, and Zhu(2016{\natexlab{b}})}]{xiao2016transg}
Xiao, H.; Huang, M.; and Zhu, X. 2016{\natexlab{b}}.
\newblock TransG: A generative model for knowledge graph embedding.
\newblock In \emph{ACL}, 2316--2325.

\bibitem[{Yang et~al.(2015)Yang, Yih, He, Gao, and Deng}]{yang2014embedding}
Yang, B.; Yih, W.; He, X.; Gao, J.; and Deng, L. 2015.
\newblock Embedding Entities and Relations for Learning and Inference in
  Knowledge Bases.
\newblock In \emph{ICLR}, 1--13.

\bibitem[{Yu et~al.(2021)Yu, Jin, Liu, He, Wang, Tong, and Han}]{AS_GCN}
Yu, Z.; Jin, D.; Liu, Z.; He, D.; Wang, X.; Tong, H.; and Han, J. 2021.
\newblock AS-GCN: Adaptive Semantic Architecture of Graph Convolutional
  Networks for Text-Rich Networks.
\newblock In \emph{ICDM}.

\bibitem[{Zhang et~al.(2019)Zhang, Tay, Yao, and Liu}]{zhang2019quaternion}
Zhang, S.; Tay, Y.; Yao, L.; and Liu, Q. 2019.
\newblock Quaternion knowledge graph embeddings.
\newblock In \emph{NeurIPS}, 2731--2741.

\end{thebibliography}
\newpage
\clearpage

\newpage
\appendix
\section{Proof for Transformations of the Special Euclidean Group}
We take a two-dimensional space as an example and
the conclusion for
higher dimensional space can be analogized accordingly.

\noindent\textbf{Proof of reflection transformation.} Denote $\theta$ as the angle of reflection. The reflection matrix can represent the reflection transformation, which is defined as follows:
\begin{equation*}
  \bm{M}_{Ref}=\left[
        \begin{array}{cc}
          \cos(\theta) & \sin(\theta) \\
          \sin(\theta) & -\cos(\theta) \\
        \end{array}
      \right].
\end{equation*}

Then we have $\bm{M}_{Ref}^{T} \bm{M}_{Ref}=\mathbf{I}$ and $\operatorname{det} \bm{M}_{Ref}=1$. It implies $\bm{M}_{Ref} \in \mathbf{S O}(n)$, so $\bm{M}_{Ref} \in \mathbf{S E}(n)$ hlods. Then the proof completes.

\noindent\textbf{Proof of inversion transformation.} According to Eq.(\ref{special orthogonal group}), if a matrix $\bm{A}\in \mathbf{S O}(n) \in \mathbf{S E}(n)$, we have $\bm{A}^T\bm{A}=\mathbf{I}$ , then we can deduce $\bm{A}$ is an inversion matrix. Therefore it can represent the inversion transformation and the proof completes.

\noindent\textbf{Proof of  translation transformation.} According to Eq.(\ref{special Euclidean group}), if a matrix $\mathbf{A}=\left[\begin{array}{cc}
\mathbf{R} & \mathbf{v} \\
0 & 1
\end{array}\right]\in \mathbf{S E}(n)$, we just make $\mathbf{R}$ be $\mathbf{0}$, then $\bm{A}$ is a translation matrix. The proof completes.

\noindent\textbf{Proof of  rotation transformation.} Denote $\theta$ as the angle of rotation. The reflection matrix can represent the rotation transformation, which is defined as follows:
\begin{equation*}
  \bm{M}_{Rot}=\left[
        \begin{array}{cc}
          \cos(\theta) & -\sin(\theta) \\
          \sin(\theta) & \cos(\theta) \\
        \end{array}
      \right].
\end{equation*}

Then we have $\bm{M}_{Rot}^{T} \bm{M}_{Rot}=\mathbf{I}$ and $\operatorname{det} \bm{M}_{Rot}=1$. It implies $\bm{M}_{Rot} \in \mathbf{S O}(n)$, so $\bm{M}_{Rot} \in \mathbf{S E}(n)$ hlods. Then the proof completes.

\noindent\textbf{Proof of  homothety transformation.} According to Eq.(\ref{special Euclidean group}), if a matrix $\bm{A} \in \mathbf{S E}(n)$, we can turn it into a diagonal matrix, then it is a homothety matrix about itself. The proof completes.

\section{Proof of Key Patterns}\label{Proof of Key Patterns}
As shown in Eq.(\ref{special orthogonal group}), to simplify the notations, we take 2-dimensional space as an example, in which all
involved embeddings are real. The conclusion in n-dimensional space can be analogized accordingly.
For a given matrix $\bm{M}=\left[
                     \begin{array}{cc}
                       a & b \\
                       c & d \\
                     \end{array}
                   \right]$, we set $\overline{\bm{M}}=\left[
                     \begin{array}{cc}
                       \frac{a}{\sqrt{a^2+b^2}} & \frac{b}{\sqrt{a^2+b^2}}  \\
                       \frac{c}{\sqrt{c^2+d^2}}& \frac{d}{\sqrt{c^2+d^2}} \\
                     \end{array}
                   \right]$.
If $\overline{\bm{M}}\in \mathbf{G} \mathbf{L}(2, \mathbb{R})$, we have $ab+cd=0$. 
We might as well let $\overline{\bm{M}}$  satisfy the conditions of $\mathbf{G} \mathbf{L}(2, \mathbb{R})$ and denote it as $\bm{R}$.
Then according to Eq.(\ref{special Euclidean group}), we have the  $\mathbf{A}=\left[\begin{array}{cc}
\mathbf{R} & \mathbf{t} \\
0 & 1
\end{array}\right]$, where $\bm{t}\in \mathbb{R}^2$.  The inverse of $\bm{A}$ is $\bm{A}^{-1}=\left[
                                                                                      \begin{array}{cc}
                                                                                        \bm{R}^{-1} & -\bm{R}^{-1}\bm{t} \\
                                                                                        0 & 1 \\
                                                                                      \end{array}
                                                                                    \right]
$. Note that since $\bm{R}\in \mathbf{G} \mathbf{L}(2, \mathbb{R})$, we have $\bm{R}^{-1}=\bm{R}^{\top}$, which means that the the reverse of $\bm{R}$ is always existing so the the reverse of $\bm{A}$ is always existing.

We denote $\otimes$ as the the  multiplication in tangent space. Next we will prove the seven patterns hold in  tangent space as follows.

\noindent\textbf{Proof of symmetry pattern.}  To prove the symmetry  pattern, if $\bm{r}(\bm{h},\bm{t})$ and $\bm{r}(\bm{t},\bm{h})$ hold, we have the equality as follows:
\begin{equation*}
  (\bm{t}=\bm{r}\otimes\bm{h})\wedge (\bm{h}=\bm{r}\otimes\bm{t})\Rightarrow \bm{r}\otimes\bm{r}=\bm{1}.
\end{equation*}

\noindent\textbf{Proof of antisymmetry pattern.}  In order to prove the antisymmetry pattern, we need to prove the
following inequality when $\bm{r}(\bm{h},\bm{t})$ and $\neg\bm{r}(\bm{t},\bm{h})$ hold:
\begin{equation*}
  (\bm{t}=\bm{r}\otimes\bm{h})\wedge (\bm{h}\neq\bm{r}\otimes\bm{t})\Rightarrow \bm{r}\otimes\bm{r}\neq\bm{1}.
\end{equation*}

\noindent\textbf{Proof of inversion pattern.} To prove the inversion pattern, if $\bm{r}_1(\bm{h},\bm{t})$ and $\bm{r}_2(\bm{t},\bm{h})$ hold, we have the equality as follows:
\begin{equation*}
  (\bm{t}=\bm{r}_1\otimes\bm{h})\wedge (\bm{h}=\bm{r}_2\otimes\bm{t})\Rightarrow \bm{r}_1=\bm{r}_2^{-1}.
\end{equation*}

\noindent\textbf{Proof of composition pattern.}  In order to prove the antisymmetry pattern, we need to prove the
following equality holds when $\bm{r}_1(\bm{h},\bm{p})$, $\bm{r}_2(\bm{h},\bm{t})$ and $\bm{r}_3(\bm{t},\bm{p})$ hold:
\begin{equation*}
  (\bm{p}=\bm{r}_1\otimes\bm{h})\wedge (\bm{t}=\bm{r}_2\otimes\bm{h}) \wedge  (\bm{p}=\bm{r}_3\otimes\bm{t})  \Rightarrow \bm{r}_1=\bm{r}_2\otimes\bm{r}_3.
\end{equation*}

\noindent\textbf{Proof of hierarchy pattern.} First of all, the hyperbolic space used by our model can capture the hierarchical structure of the data. Next we prove that the model can also logically derive the hierarchy pattern. To prove the hierarchy pattern, if $\bm{r}_1(\bm{h},\bm{t})$ and $\bm{r}_1(\bm{h},\bm{t})$ hold, we need to prove  the following equality holds:
\begin{equation*}
\bm{r}_1(\bm{h},\bm{t})\Rightarrow \bm{r}_2(\bm{h},\bm{t}).
\end{equation*}
Then we need to prove
\begin{equation}\label{hierarchy pattern}
\bm{t}=\bm{r}_1\otimes\bm{h}\Rightarrow \bm{t}=\bm{r}_2\otimes\bm{h}.
\end{equation}

We  assume that the Eq.(\ref{hierarchy pattern}) holds, then we have
\begin{equation}\label{hierarchy pattern wedge}
  (\bm{t}=\bm{r}_1\otimes\bm{h})\wedge (\bm{t}=\bm{r}_2\otimes\bm{h}) \Rightarrow \bm{r}_1\otimes\bm{h}=\bm{r}_2\otimes\bm{h}.
\end{equation}

we denote $\bm{r}_1$ as $\left[\begin{array}{cc}
\mathbf{R}_1 & \mathbf{t}_1 \\
0 & 1
\end{array}\right]$ and $\bm{r}_2$ as $\left[\begin{array}{cc}
\mathbf{R}_2 & \mathbf{t}_2 \\
0 & 1
\end{array}\right]$, where $\bm{R}_1=\left[
                     \begin{array}{cc}
                       \frac{a_1}{\sqrt{a_1^2+b_1^2}} & \frac{b_1}{\sqrt{a_1^2+b_1^2}}  \\
                       \frac{c_1}{\sqrt{c_1^2+d_1^2}}& \frac{d_1}{\sqrt{c_1^2+d_1^2}} \\
                     \end{array}
                   \right]$ with $a_1b_1+c_1d_1=0$, $\bm{R}_2=\left[
                     \begin{array}{cc}
                       \frac{a_2}{\sqrt{a_2^2+b_2^2}} & \frac{b_2}{\sqrt{a_2^2+b_2^2}}  \\
                       \frac{c_2}{\sqrt{c_2^2+d_2^2}}& \frac{d_2}{\sqrt{c_2^2+d_2^2}} \\
                     \end{array}
                   \right]$ with $a_2b_2+c_2d_2=0$, $\bm{t}_1 \in \mathbb{R}^2$ and $\bm{t}_2 \in \mathbb{R}^2$. Then we bring $\bm{r}_1$ and $\bm{r}_2$ into the Eq.(\ref{hierarchy pattern wedge}). Such we can get the solutions of $\bm{r}_1$ and  $\bm{r}_2$ after solving the equation,  which can satisfy the Eq.(\ref{hierarchy pattern}). Then we have $\bm{r}_1(\bm{x},\bm{y})\Rightarrow \bm{r}_2(\bm{x},\bm{y})$.

\noindent\textbf{Proof of intersection pattern.} In order to prove the intersection pattern, if $\bm{r}_1(\bm{h},\bm{t})$, $\bm{r}_2(\bm{h},\bm{t})$ and $\bm{r}_3(\bm{h},\bm{t})$ hold, we
have the following equality:
\begin{equation*}
\begin{aligned}
  &(\bm{t}=\bm{r}_1\otimes\bm{h})\wedge (\bm{t}=\bm{r}_2\otimes\bm{h})
   \Rightarrow (\left[\begin{array}{cc}
\mathbf{R}_1 & \mathbf{v}_1 \\
0 & 1
\end{array}\right]\otimes \left[
                                  \begin{array}{c}
                                    \bm{h} \\
                                    1 \\
                                  \end{array}
                                \right]\\
=&\left[
                                  \begin{array}{c}
                                    \bm{t} \\
                                    1 \\
                                  \end{array}
                                \right]) \wedge \left(\left[\begin{array}{cc}
\mathbf{R}_2 & \mathbf{v}_2 \\
0 & 1
\end{array}\right]\otimes \left[
                                  \begin{array}{c}
                                    \bm{h} \\
                                    1 \\
                                  \end{array}
                                \right]=\left[
                                  \begin{array}{c}
                                    \bm{t} \\
                                    1 \\
                                  \end{array}
                                \right]\right).
\end{aligned}
\end{equation*}
Then we have
\begin{equation*}
  \bm{R}_1\otimes\bm{h}+\bm{v}_1=\bm{t}, \ \bm{R}_2\otimes\bm{h}+\bm{v}_2=\bm{t}.
\end{equation*}

We multiply the left and right parts of the first equation by $\bm{v}_2$ respectively,  and we multiply the left and right parts of the second equation by $\bm{v}_1$ respectively. Then we have:
\begin{equation*}
  \bm{v}_2\otimes\bm{R}_1\otimes\bm{h}+\bm{v}_2\otimes\bm{v}_1=\bm{v}_2\otimes\bm{t}, \ \bm{v}_1\otimes\bm{R}_2\otimes\bm{h}+\bm{v}_1\otimes\bm{v}_2=\bm{v}_1\otimes\bm{t}.
\end{equation*}
Make the first formula subtract the second formula above, and we can get
\begin{equation*}
  (\bm{v}_2\otimes\bm{R}_1-\bm{v}_1\otimes\bm{R}_2)\otimes\bm{h}=(\bm{v}_2-\bm{v}_1)\otimes\bm{t}.
\end{equation*}
Since $\bm{r}_1\neq \bm{r}_2$, $v_1\neq v_2$. Then we have \begin{equation*}
  (\bm{v}_2-\bm{v}_1)^{-1}\otimes(\bm{v}_2\otimes\bm{R}_1-\bm{v}_1\otimes\bm{R}_2)\otimes\bm{h}=\bm{t}.
\end{equation*}
Next let $\bm{r}_3=\left[\begin{array}{cc}
\mathbf{R}_3 & \bm{v}_3 \\
0 & 1
\end{array}\right]$, where $\mathbf{R}_3=  (\bm{v}_2-\bm{v}_1)^{-1}\otimes(\bm{v}_2\otimes\bm{R}_1-\bm{v}_1\otimes\bm{R}_2)$ and $\bm{v}_3=\bm{0}$.
Then we have $\bm{r}_1(\bm{x},\bm{y}) \wedge \bm{r}_2(\bm{x},\bm{y}) \Rightarrow\bm{r}_3(\bm{x},\bm{y})$.

\noindent\textbf{Proof of mutual exclusion pattern.} To prove the mutual exclusion pattern, if $\bm{r}_1(\bm{h},\bm{t})$ and $\bm{r}_2(\bm{h},\bm{t})$ hold, where $\bm{r}_1$ and $\bm{r}_2$ are orthogonal ($\bm{r}_1$$\bm{r}_2=\bm{0}$),
we have the formula as follows:
\begin{equation*}
\begin{aligned}
  &(\bm{t}=\bm{r}_1\otimes\bm{h})\wedge (\bm{t}=\bm{r}_2\otimes\bm{h})\\
  \Rightarrow& \bm{r}_1\otimes\bm{h}=\bm{r}_2\otimes\bm{h}\\
  \Rightarrow& \bm{r}_1\otimes\bm{r}_1\otimes\bm{h}=\bm{r}_1\otimes\bm{r}_2\otimes\bm{h}\\
  \Rightarrow& \bm{r}_1\otimes\bm{r}_1\otimes\bm{h}=\bm{0}\\
  \Rightarrow&\perp.
  \end{aligned}
\end{equation*}
Therefore, we have $\bm{r}_1(\bm{x},\bm{y})\wedge \bm{r}_2(\bm{x},\bm{y})\Rightarrow \perp$.

\section{Hierarchy Estimates}\label{Hierarchy estimates}There are two main metrics to estimate how hierarchical a
relation is, which is called  the curvature estimate $\xi_G$ (it means how much the graph is tree-like) and the Krackhardt hierarchy score Khs$_G$ (it means how many small loops in the graph). Each of these two measures has its own merits: the  global hierarchical behaviours can be captured by the curvature
estimate, 
and the local
behaviour can be captured by the Krackhardt score.  

\noindent\textbf{Curvature estimate.} In order to estimate the curvature
of a relation $r$, we restrict to the undirected graph
$G_r$ spanned by the edges labeled as $r$. As  shown in  \cite{DBLP:conf/iclr/GuSGR19}, for the given vertices $\{a,b,c\}$, we denote $\xi_{G_r}(a,b,c)$ as the curvature
estimate of a triangle in $G_r$, which can be  given as follows:
\begin{equation*}
\begin{aligned}
\xi_{G_{r}}(a, b, c) =&\frac{1}{2 d_{G_{r}}(a, m)}\left(d_{G_{r}}(a, m)^{2}
+d_{G_{r}}(b, c)^{2} / 4\right.\\
&-\left.\left(d_{G_{r}}(a, b)^{2}+d_{G_{r}}(a, c)^{2}\right) / 2\right),
\end{aligned}
\end{equation*}
 where $m$ denotes the midpoint of the shortest path from $b$ to $c$.
 For the triangle,  its  curvature estimate  is positive if it is in circles;  its  curvature estimate  is zero if it is in lines;  its  curvature estimate  is negative if it is in trees.
  Moreover, following \cite{DBLP:conf/iclr/GuSGR19},  suppose there is a triangle in a Riemannian manifold $M$ lying on a plane, and we can use $\xi_M(a,b,c)$ to estimate
the sectional curvature of the plane.
 Denote  the total number of connected
components in $G_r$ as $m_r$, denote  the number of nodes
in the component $c_{i,r}$ as $N_{i,r}$ and denote  the mean of the estimated curvatures of the sampled triangles as $\xi_{G_r}$.  We then sample 1000 $w_{i,r}$ triangles
from each connected component $c_{i,r}$ of $G_r$ where
$w_{i,r}=\frac{N_{i,r}^3}{\sum_{i=1}^{m_r}N_{i,r}^3}$.
Next, we  take the weighted average of the
relation curvatures $\xi_{G_r}$ with respect to the weights
$\frac{\sum_{i=1}^{m_r}N_{i,r}^3}{\sum_r\sum_{i=1}^{m_r}N_{i,r}^3}$ for
all the  graphs.

\noindent\textbf{Krackhardt hierarchy score.} For the directed
graph $G_r$  which is spanned by the relation $r$, we denote
the adjacency matrix ( If there is an edge
connecting node $i$ to node $j$, we have $R_{i,j} = 1$. If there is not an edge
connecting node $i$ to node $j$, we have $R_{i,j} \neq 1$.) as
$R$. Then we have:
\begin{equation}
\mathrm{Khs}_{G_{r}}=\frac{\sum_{i, j=1}^{n} R_{i, j}\left(1-R_{j, i}\right)}{\sum_{i, j=1}^{n} R_{i, j}}
\end{equation}
More details can refer to \cite{DBLP:journals/socnet/EverettK12}. For fully observed symmetric relations, each edge of which is in a two-edge loop, it can be seen  that  Khs$_{G_r} = 0$.
And for the anti-symmetric relations,
we have Khs$_{G_r} = 1$.

\begin{table*}
\begin{center} 
 \begin{tabular}{lcccccccc}
  \toprule
relation name& $\xi_{G}$& Khs& MuRE &MuRP &MuRS &MuRMP &MuRMP-a& GIE\\
  \midrule
 also$\_$see& -2.09& .24  &.634 &.705& .483 &.692 &.725&$\bm{.759}$\\
hypernym &-2.46 &.99 &.161 &.228 &.126 &.222 &.232&$\bm{.262}$\\
has$\_$part& -1.43& 1 &.215 &.282 &.301& .134 &.316&$\bm{.334}$\\
member$\_$meronym &-2.90 &1 &.272 &.346 &.138 &.343 &$.350$&$\bm{.360}$\\ 
synset$\_$domain$\_$topic$\_$of& -0.69 &.99 &.316& .430& .163& .421 &$\bm{.445}$&.435\\ 
instance$\_$hypernum& -0.82& 1 &.488 &.471 &.258 &.345 &.491&$\bm{.501}$\\ 
member$\_$of$\_$domain$\_$region& -0.78& 1 &.308& .347& .201& .344& .349&$\bm{.404}$\\ 
member$\_$of$\_$domain$\_$usage& -0.74 & 1 &.396 &.417 &.228 &.416 &$.420$&$\bm{.438}$\\  
derivationally$\_$related$\_$form &-3.84 &0.4 &.954 &.967 &.965 &.967 &$\bm{.970}$&$.968$\\
similar$\_$to& -1.00 &0 &$\bm{1}$ &$\bm{1}$ &$\bm{1}$ &$\bm{1}$& $\bm{1}$&$\bm{1}$\\
verb$\_$group &-0.5& 0 &.974& .974 &.976 &.976& .981&$\bm{.984}$\\
  \bottomrule
 \end{tabular}
 \caption{\label{11 relations} Comparison of hits@10 for WN18RR. 
MuRMP-a represents MuRMP-autoKT in \cite{wang2021mixed}.
}
 \end{center}
\end{table*}

\begin{table*}
\begin{center}
 \begin{tabular}{c|cccc|cccc|cccc}
  \toprule
  $\ $ &  \multicolumn{4}{c}{WN18RR}  &\multicolumn{4}{c}{FB15k-237}&\multicolumn{4}{c}{YAGO3-10} \\
  Analysis  &MRR &H@1 &H@3 &H@10  &MRR &H@1 &H@3 &H@10&MRR &H@1 &H@3 &H@10\\
  \midrule
 GIE$_1$ &.472 &.438 &.474 &.534 &.336 &.231 &.373 &.526 &.534 &.477 &.584 &.680\\
  GIE$_2$ &.477 &.440 &.483 &.541 &.338 &.236 &.371 &.526 &.538 &.479 &.588 &.674\\
   GIE$_3$ &.471 &.436 &.470 &.531 &.332 &.228 &.375 &.523 &.531 &.473 &.581 &.671\\
  \bottomrule
 \end{tabular}
 \caption{ Analysis on different variants of scoring function.}\label{Ablation Study on Score Function}
 \end{center}
\end{table*}

\section{Experimental Details}

 We
conduct the hyperparameter search for the negative sample size, learning
rate, batch size, dimension and the number of epoch. We train each model for 150 epochs and use early stopping after 100 epochs if the validation MRR stops increasing.
 Moreover, we  count the running time of each epoch of GIE on different datasets: 20s for WN18RR, 34s for FB15K237, 65s for YAGO3-10. We need to run 150 epochs in total. Therefore, the overall time for each datasets is between 50 minutes and 162 minutes.

In the test dataset, we  measure and compare the performance  of models listed in the paper  for each test
triplet $(h,r,t)$. For all $h' \in \mathcal{E}$ (where $\mathcal{E}$ is the set of entities),   we calculate the scores of  triplets $(h',r,t)$ and then denote the ranking
 of the triplet $(h,r,t)$ among these triplets as $rank_h$. Next, for all $t'\in \mathcal{E}$, we compute the scores of  triplets $(h,r,t')$ and denote
the ranking  of the triplet $(h,r,t)$ among these triplets as $rank_t$. According to the calculation above, MRR is calculated as follows, which is the mean of the reciprocal rank for the  test dataset:
\begin{equation*}
  MRR=\frac{1}{2*|\Omega|}\sum_{(h,r,t)\in \Omega }\frac{1}{rank_h(h,r,t)}+\frac{1}{rank_t(h,r,t)}.
\end{equation*}
where $\Omega$ denotes the set of observed triplets in the test dataset.

\end{document}